\documentclass[10pt,twocolumn]{article}

\usepackage[margin=0.72in]{geometry}
\usepackage[T1]{fontenc}
\usepackage[utf8]{inputenc}
\usepackage{graphicx}
\usepackage{amsmath,amssymb}
\usepackage{booktabs}
\usepackage{array}
\usepackage{tabularx}
\usepackage{tikz}
\usepackage{url}
\usepackage[hidelinks,hypertexnames=false]{hyperref}
\usepackage[protrusion=true,expansion=false]{microtype}
\usepackage{cuted}
\usepackage{float}
\usetikzlibrary{positioning,arrows.meta,calc}

\title{\vspace{-1.2em}Win by Silence:\\Deletion Non-Monotonicity, Autonomous Exploitation, and Typed-State Gating in LLM Plan Evaluation}
\author{Aleh Manchuliantsau\\Independent Researcher\\\texttt{aleh.manchuliantsau@gmail.com}}
\date{Version 1.0: July 13, 2026}

\newcommand{\ci}[2]{95\% CI [#1, #2]}
\newcommand{\usd}[1]{\$#1}

\begin{document}
\maketitle

\begin{abstract}
Plan evaluators can reward a strategic plan for becoming less explicit. This paper studies this failure in a staged expected-value scorer for LLM-generated venture routes. Proposition~1 gives the deletion score change when an interior transition is removed, the predecessor is retargeted, and downstream continuation value is retained:
\(\Delta_k=(\prod_{i<k}p_i)[c_k+(1-p_k)R_{k+1}]\).
On the frozen 26-route cohort, every one of 57 admissible deletions matched the analytic identity, every sign matched the corollary threshold, and every route contained at least one score-improving admissible deletion. A score-seeking optimizer, allowed to restructure routes but not told that removing a transition could increase score or create an uncovered delta, discovered baseline-beating uncovered structures in 21/26 routes. On the frozen splice benchmark, GATE refused score release for 26/26 silenced routes with 0/26 honest suspensions; under GATE feedback, 47/54 post-refusal revisions repaired to a covered structure, and strict covered improvement rose from 1/26 to 13/26. Finally, an adaptive compiler-aware co-author exposed the registry-provenance boundary: obligation-channel evasions remained 6/6 in all four v1/v1.5 conditions, while delta-indexed cost floors narrowed the score channel, reducing beat-honest routes from 6/6 to 3/6 and fundability-by-silence from 5/6 to 0/6 without establishing semantic completeness. If a plan's score improves only because it stopped mentioning necessary work, the plan did not improve---the evaluation created an omission incentive. PCSC detects and neutralizes post-hoc omission splices given model-mediated typed-state records. In the cooperative setting tested, GATE therefore functions as a deterministic search-shaping constraint rather than only a post-hoc filter. It does not verify the semantic completeness or real-world quality of arbitrary LLM-generated strategies.
\end{abstract}

\section{Introduction}

LLM-generated strategic plans are increasingly scored, selected, and iteratively optimized before the world has supplied outcomes. A route can be made more persuasive by editing prose, but it can also be made more attractive to a scorer by becoming less explicit about necessary work. In a staged evaluator, a shorter route may carry fewer costs and fewer probability discounts while still claiming the same terminal value. The problem is not merely imperfect judging. It is value being credited across unpaid state changes.

This paper studies that omission incentive in the RouteCast/PCSC experimental rail, which follows Paper~1's code-owned delayed-ground-truth setting but asks a narrower integrity question \cite{manchuliantsau2026routecast}. A route is a Wedge--Bridge--Vision chain. Each transition carries a probability \(p_i\), a cost \(c_i\), and typed fields describing the work that moves one state to the next. The shipped scorer folds costs and terminal value through the chain. The vulnerability is that an edited plan can splice out a load-bearing transition, retarget the predecessor to the deleted transition's target, and still receive downstream value.

I use \emph{Win-by-Silence} to denote score improvement by omission while terminal value is retained. An \emph{uncovered delta} is a claimed state change without a discharged obligation covering that change.

The central claim is one causal argument, not a bundle of independent experiments:
\begin{enumerate}
\item Proposition~1 characterizes the deletion reward.
\item The reward exists across the frozen route cohort.
\item A score-seeking optimizer finds uncovered-delta improvements without being told the exploit mechanism or which edits would trigger it.
\item A typed-state coverage gate redirects optimization toward covered structures by refusing score release over uncovered seams.
\item Adaptive co-authoring exposes the model-mediated registry provenance limit; delta-indexed floors close much of the score channel but do not establish semantic completeness.
\end{enumerate}

Section~2.4 states the PCSC interface used here. A proof obligation is an implementation-defined predicate, not a philosophical, legal, or real-world proof. PCSC denotes a proof-carrying strategy compiler over these implemented predicates, not a formal proof system for real-world strategic correctness.

The contribution hierarchy is therefore led by the deletion characterization. The mutation map, optimizer test, deterministic typed-state gate, value coupling, and adaptive-boundary experiment are evidence and consequences of that mechanism. All results are benchmark- and scorer-specific. The words ``complete,'' ``verification,'' and ``proof'' refer only to implemented predicates over the frozen records; no result establishes prospective decision utility or completeness of arbitrary LLM-generated strategies.

\section{Setting and Scope}

\subsection{Route Records and Staged Scores}

Let a route be a connected zero-indexed chain of transitions \(e_0,\ldots,e_{n-1}\) from an as-of state to a terminal Vision state. Transition \(e_i\) has a scorer-derived probability \(p_i\in[0,1]\) and cost \(c_i\geq 0\). The terminal state after \(e_{n-1}\) has a value \(V\). The staged score is the backward fold
\[
R_i=-c_i+p_iR_{i+1},\qquad R_n=V .
\]
This is a standard staged expected-value or real-options-style recursion \cite{dixit1994investment,gompers1995optimal}. No novelty is claimed for staged investment math.

The score is not real-world plan quality. It is a route score under one frozen scorer. In Paper~1, the scorer was used as a provisional forecast-ranking under delayed ground truth, with point-in-time evidence packets and code-owned arithmetic \cite{manchuliantsau2026routecast}. This paper asks whether that scorer can be made to reward omission.

\subsection{Admissible Deletion}

An admissible interior deletion selects transition \(e_k\) with \(k\in[1,n-2]\): the first as-of transition \(e_0\) and terminal Vision-reaching transition \(e_{n-1}\) are excluded. The mutation deletes \(e_k\) and retargets the predecessor to the deleted transition's target. The predecessor keeps its own \(p\) and \(c\). The terminal value and downstream suffix are retained. Fixed-parameter deletion holds the original \(p,c,V\) quantities fixed except for the deleted stage; end-to-end pipeline re-derivation reruns the route through the scorer and may change transition costs and probabilities over the spliced graph. The algebra below concerns the fixed-parameter fold; both channels are reported empirically.

An uncovered typed-state delta is a target state change that is claimed by the record but not covered by the transition's discharged obligations. A discharged status means the implemented predicate passes; undischarged means the delta is unpaid under that implementation.

\subsection{Model-Mediated Typing}

PCSC v1 constructs typed state records from authored plans and then performs deterministic checks over those records. Registry provenance is model-mediated typing: the state fields are derived from model-authored route text. The deterministic result concerns post-hoc splices applied to those typed records. Adversarially co-authored consistent states remain outside the v1 guarantee, and Section~\ref{sec:adaptive} measures that boundary.

\subsection{PCSC Interface: GATE and DOCK}

PCSC treats a route as a sequence of typed source-state to target-state transitions. Each claimed typed state delta is associated with an implementation-defined set of proof obligations: decidable predicates over the transition, its states, required fields, or the claimed state delta itself. An obligation is discharged when its predicate passes. A material seam or delta is uncovered when the record claims a material state change but the required obligation set is not discharged.

The protocol invariant is that no strategic state change may receive score or terminal-value credit without a discharged obligation covering that delta. GATE enforces score admissibility: when a material uncovered seam is present, it refuses score release and returns a suspend-and-ask outcome together with the undischarged-obligation list. DOCK instead enforces value admissibility by removing terminal-value credit over an uncovered path before releasing the resulting score.

These rules define an interface over typed records rather than a claim of real-world proof. State typing is model-mediated, while the subsequent checks are deterministic. Passing GATE or DOCK therefore establishes only compliance with the implemented predicates over the supplied record. It does not establish semantic completeness, strategic correctness, prospective decision utility, or completeness of arbitrary LLM-generated plans.

\section{Deletion Non-Monotonicity}

\textbf{Proposition 1.} For the staged fold \(R_i=-c_i+p_iR_{i+1}\), deleting admissible interior transition \(e_k\) while retargeting the predecessor and retaining \(R_{k+1}\) changes the score by
\[
\Delta_k =
\left(\prod_{i<k} p_i\right)
\left[c_k + (1-p_k)R_{k+1}\right].
\]
For \(\prod_{i<k}p_i>0\) and \(p_k<1\),
\[
\Delta_k>0
\iff
R_{k+1}>-\frac{c_k}{1-p_k}.
\]

\textbf{Proof.} Before deletion, the contribution at \(k\) is \(R_k=-c_k+p_kR_{k+1}\). After deletion, the predecessor points directly to \(R_{k+1}\). Thus the local suffix change is
\[
R_{k+1}-R_k
=R_{k+1}-(-c_k+p_kR_{k+1})
=c_k+(1-p_k)R_{k+1}.
\]
Multiplying by the probability prefix \(\prod_{i<k}p_i\) gives \(\Delta_k\), and rearranging the positive-delta condition under the stated hypotheses gives the corollary. \(\square\)

The identity separates a removed-cost channel,
\[
\left(\prod_{i<k}p_i\right)c_k,
\]
from a retained-value channel,
\[
\left(\prod_{i<k}p_i\right)(1-p_k)R_{k+1}.
\]
If the continuation value is positive or not too negative, the scorer can reward deletion. If the continuation is costly and negative, an early deletion can hurt because it removes a probability factor that had been discounting downstream costs.

\begin{strip}
\begin{center}
\small
\makebox[\textwidth][c]{%
\begin{tikzpicture}[
  state/.style={draw, rounded corners=2pt, minimum width=2.1cm, minimum height=.68cm, align=center, fill=gray!5},
  edge/.style={-{Latex[length=2mm,width=1.45mm]}, line width=.55pt},
  splice/.style={-{Latex[length=2mm,width=1.45mm]}, line width=.7pt},
  ghost/.style={-{Latex[length=1.35mm,width=1mm]}, dashed, line width=.32pt, draw=black!55},
  gone/.style={draw, dashed, rounded corners=2pt, minimum width=2.1cm, minimum height=.78cm, align=center, fill=gray!7},
  lab/.style={font=\scriptsize, align=center}
]
\node[state] (s0) at (0,0) {as-of\\state};
\node[state] (s1) at (4,0) {intermediate\\state};
\node[state] (s2) at (8,0) {target\\state};
\node[state] (sv) at (12,0) {Vision\\value \(V\)};
\coordinate (spancenter) at ($(s1)!0.5!(s2)$);

\draw[edge] (s0) -- node[above, lab] {\(p_0,c_0\)} (s1);
\draw[splice] (s1) -- (s2);
\node[lab, anchor=south] at ($(spancenter)+(0,.44)$) {retargeted predecessor\\claims farther state};
\draw[edge] (s2) -- node[above, lab] {suffix\\\(R_{k+1}\)} (sv);

\node[gone] (sk) at ($(spancenter)+(0,-1.55)$) {deleted edge \(e_k\)\\[-1pt]{\scriptsize \(p_k,c_k\)}};
\draw[ghost] (s1.south) -- (sk.north west);
\draw[ghost] (sk.north east) -- (s2.south);
\node[lab, below=.15cm of sk] {original path removed};

\node[lab, text width=5.7cm] (delta) at ($(spancenter)+(0,-3.0)$) {uncovered typed-state delta:\\the plan stopped mentioning necessary work\\while terminal value remains credited};
\node[draw, rounded corners=2pt, align=left, anchor=north, fill=white, text width=7.25cm] (formula) at ($(delta.south)+(0,-.55cm)$) {
\(\Delta_k=\left(\prod_{i<k}p_i\right)\left[c_k+(1-p_k)R_{k+1}\right]\)\\[2pt]
removed-cost channel: \(\left(\prod_{i<k}p_i\right)c_k\)\\
retained-value channel: \(\left(\prod_{i<k}p_i\right)(1-p_k)R_{k+1}\)
};
\end{tikzpicture}
}
\par\vspace{.6em}
\refstepcounter{figure}
\label{fig:wbs-schematic}
\begin{minipage}{\textwidth}
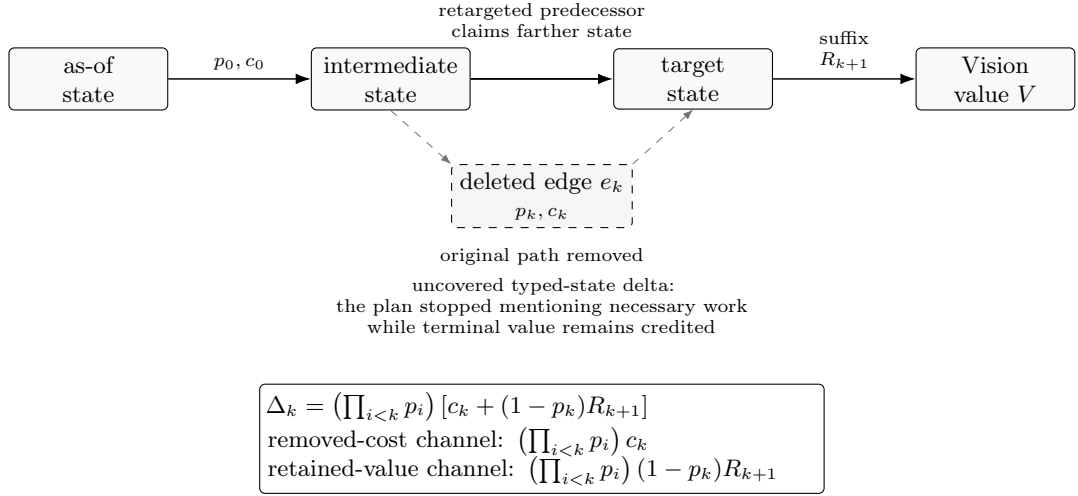
Figure~\thefigure: Win-by-Silence splice. An interior transition is removed from the plan record, the predecessor is retargeted, and downstream value is retained. The score gain decomposes into a removed-cost channel and a retained-value channel.
\end{minipage}
\end{center}
\vspace{.8em}

\end{strip}

In the searched corpus, I found no earlier explicit closed-form characterization of when deleting an interior stage increases the value assigned by this staged expected-value fold while the downstream continuation value is retained. The broader phenomenon is known in other forms: plan-reduction work removes redundant actions while preserving validity and reducing cost \cite{nakhost2010action,salerno2025minimal}; process-model repair introduces skip transitions while preserving final markings and improving alignment fitness \cite{polyvyanyy2016impact}; reward-hacking work documents reward obtained by omitting intended work \cite{murphy2013first,amodei2016concrete,krakovna2020specification,macdiarmid2025natural,thaman2026rhb,zhao2026specbench}; LLM judges can gap-fill omitted reasoning \cite{wu2026logicgraph}; and declarative conformance can count vacuous satisfaction \cite{diciccio2018relevance}. The distinction here is the record-level, value-retaining identity for a plan valuation fold and the coupling of its remedy to value admissibility.

\section{Experimental Program}

All quantitative statements below are reconstructed from frozen artifacts under \texttt{eval/pcsc/}. Hash sidecars verified with \texttt{shasum -a 256 -c eval/pcsc/*.sha256}. Binary intervals are Wilson 95\% intervals. Arms, detectors, and deterministic versus model-mediated outputs are never pooled.

\begin{table*}[t]
\centering
\scriptsize
\caption{Experimental sequence. ``Model calls'' counts model-mediated generation or judging calls in that phase; deterministic analyses are zero-model.}
\label{tab:program}
\begin{tabularx}{\textwidth}{p{1.55cm}p{2.7cm}p{1.8cm}p{1.8cm}p{2.15cm}p{2.2cm}X}
\toprule
Phase & Question & Unit & Model calls & Preregistration & Model-mediated component & Main limitation \\
\midrule
Phase 0 & Does a sampled deletion improve the shipped pipeline score? & 26 routes, two arms & 26 decompositions & \texttt{PREREG.md} & route decomposition & one temperature-zero draw; pipeline re-derives quantities \\
EXP A & Does Proposition~1 explain every admissible deletion? & 57 deletions & zero & \texttt{PREREG\_expA.md} & none & fixed-parameter fold, one cohort \\
Phase 1/v1 & Can coverage obligations catch the splice? & 26 seeded splices, 111 gold edges & v0 judge only & \texttt{PREREG\_phase1.md}, \texttt{PREREG\_v1.md} & v0 judge; model-authored state typing & v1 is deterministic after typing, not arbitrary-plan completeness \\
\shortstack[l]{GATE\\DOCK} & Does coverage affect score release/value? & 26 honest/silenced pairs & zero & \texttt{coupling prereg} & typed registry inherited from v1 & DOCK misses registered relative-win collapse \\
EXP D & Does full-context judging remove the objection? & 52 routes, 26 pairs & 104 judge calls & \texttt{PREREG\_expD.md} & full-context judge & one judge family, \(K=1\) \\
EXP B & Does a score optimizer discover silence? & 26 packets, two arms, \(K=5\) & 269 optimizer calls & \texttt{PREREG\_expB.md} & optimizer outputs & same-family optimizer/gold authoring; detector registry limit \\
EXP C & What happens against compiler-aware co-authoring? & 6 packets, four conditions, \(K=6\) & 169 adversary calls & \texttt{PREREG\_v2.md} & adversary outputs; no LLM judge & small adaptive red team; semantic substantiation remains future work \\
\bottomrule
\end{tabularx}
\end{table*}

\section{Mutation Surface and Optimizer Discovery}

\subsection{Phase 0: Sampled Deletion}

Phase 0 applied two frozen splice arms to the same 26 de-identified as-of packets: a highest-cost admissible deletion and a uniformly random admissible deletion. Under the shipped end-to-end pipeline, silence scored at least as high as honest in 21/26 routes in the max-cost arm (80.8\%, \ci{62.1}{91.5}) and 21/26 in the random-control arm (80.8\%, \ci{62.1}{91.5}). Each arm produced 1/26 fundability flips (3.8\%, \ci{0.7}{18.9}). The result was pre-registered as structural only if both arms helped, and both did.

Phase 0 also produced a misleading diagnostic: five 5-edge routes appeared to resist deletion. EXP A resolves that account. Those routes resist the re-derivation pipeline's particular sampled splice, not the fixed fold itself.

\subsection{EXP A: Exhaustive Deletion Surface}

\begin{figure*}[t]
\centering
\scriptsize
\makebox[\textwidth][c]{%
\begin{minipage}[c]{.46\textwidth}
\centering
\begin{tikzpicture}[x=.052cm,y=.054cm]
  \draw[->] (-55,0) -- (50,0) node[right] {analytic \(\Delta\) (M USD)};
  \draw[->] (0,-60) -- (0,50) node[above] {observed \(\Delta\) (M USD)};
  \draw[dashed] (-55,-55) -- (50,50);
  \foreach \x/\y in {-45/-45,-30/-30,-15/-15,15/15,30/30,45/45} {
    \draw[fill=black] (\x,\y) circle (1.05pt);
  }
  \node[align=left, anchor=west] at (-52,38) {57/57 analytic-observed\\agreement; max diff\\\(1.49{\times}10^{-8}\) USD};
  \node[align=left, anchor=west] at (8,-48) {27/57 positive\\30/57 negative\\0 ties};
\end{tikzpicture}
\end{minipage}\hspace{.015\textwidth}%
\begin{minipage}[c]{.495\textwidth}
\centering
\begin{tikzpicture}[x=.258cm,y=.455cm]
  \node[anchor=west] at (0,7.4) {per-route improving seams};
  \foreach \i in {1,...,26} {
    \draw[gray!35] (\i,0) -- (\i,6);
    \fill[black] (\i,5) rectangle +(0.17,.45);
  }
  \fill[black] (22,2.8) rectangle +(0.17,.45);
  \draw[->] (0,0) -- (27.2,0) node[right] {\scriptsize route};
  \draw[->] (0,0) -- (0,6.5) node[above] {\scriptsize seam};
  \node[align=left, anchor=west] at (2,1.1) {last eligible seam\\improves in 26/26};
  \node[align=left, anchor=west] at (18,2.1) {R22 has one\\additional improving seam};
\end{tikzpicture}
\end{minipage}%
}
\caption{EXP A mutation surface. The fixed-parameter deletion identity matches the shipped staged fold on every admissible deletion. Every route has at least one improving deletion; the universal improving position is the last eligible seam.}
\label{fig:expA}
\end{figure*}
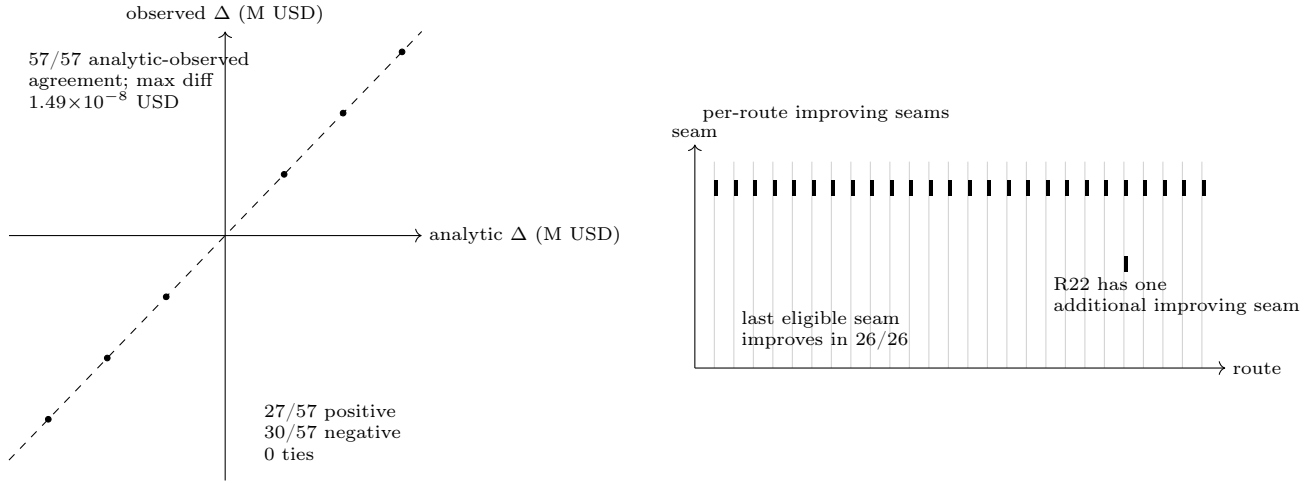

EXP A swept every admissible interior deletion of every frozen route, with zero model calls and zero network. There were 57 admissible deletions over 26 routes. P1 was MET: analytic and observed deltas agreed for 57/57 deletions (100\%, \ci{93.7}{100.0}), with maximum absolute difference \(1.49\times 10^{-8}\) USD. P3 was MET: 57/57 signs matched the corollary threshold.

P2 was NOT MET. The registered prediction was that the five Phase-0-resistant routes would be exactly the all-\(\Delta\leq 0\) routes. The actual all-nonpositive set was empty. Every route, including all five formerly ``resistant'' routes, had at least one score-improving deletion (26/26; \ci{87.1}{100.0}). The last eligible seam improved in 26/26 routes; R22 additionally improved at an earlier seam. Overall, 27/57 deletions were score-improving (47.4\%, \ci{35.0}{60.1}) and 30/57 hurt.

The reconciliation separates fold arithmetic from pipeline re-derivation. Among 26 phase-0 arms with zero \((p,c)\)-vector drift, there were 0/26 sign flips (0.0\%, \ci{0.0}{12.9}). All 16 sign flips occurred in the 26 drift arms (61.5\%, \ci{42.5}{77.6}). Thus the fold's deletion reward exists across the cohort, while pipeline re-derivation can mask or amplify it.

\subsection{EXP B Arm 1: Autonomous Discovery}

EXP A and EXP B use the same frozen 26-route cohort for different purposes. EXP A establishes that the deletion incentive exists under the registered mutation operator and fixed-parameter fold. EXP B asks whether score-seeking optimization can find behaviorally relevant uncovered-delta improvements in that same setting. This is mechanism plus behavioral relevance, not an independent prevalence replication.

The frozen optimizer prompt instructed the model to improve staged score, keep terminal Vision, and allowed restructuring, merging, adding, or removing transitions. The optimizer was therefore allowed to remove transitions. What it was not told was the exploit mechanism: that removing a transition could increase score by retaining downstream value while creating an uncovered typed-state delta.

Arm 1 used score feedback only. The optimizer discovered a baseline-beating uncovered-delta structure in 21/26 packets (80.8\%, \ci{62.1}{91.5})---coincidentally the same count as Phase 0, but a distinct measurement of optimizer discovery rather than seeded-splice performance---usually by revision 1--2, without being told which edits would trigger the exploit. Deletion or merge occurred in 74/130 applied revisions (56.9\%, \ci{48.3}{65.1}). Among score-improving revisions, 53/54 carried an uncovered typed-state delta (98.1\%, \ci{90.2}{99.7}); the detector-independent \texttt{seam\_skip} variant appeared in 9/54. Strict genuine covered improvement was only 1/26 packets (3.8\%, \ci{0.7}{18.9}).

The mutation surface shows that the incentive exists under the mutation operator; the optimizer result shows that score-seeking search can find related uncovered-delta improvements on the same cohort.

\section{Detection, Inadmissibility, and Redirection}

\subsection{Detection Ladder}

\begin{table*}[t]
\centering
\scriptsize
\caption{Detection ladder. Denominators differ by detector and unit; rows are not pooled.}
\label{tab:detection}
\begin{tabularx}{\textwidth}{p{2.5cm}p{2.4cm}p{2.6cm}p{2.6cm}p{2.2cm}X}
\toprule
Detector & Unit/context & Seam detections & Honest flags & Model-mediated? & Interpretation \\
\midrule
Structural proxy & 1C edge structure & 11/26 (42.3\%; \ci{25.5}{61.1}) & 33/111 (29.7\%; \ci{22.0}{38.8}) & no & structure alone misses many splices and over-fires \\
Blinded semantic judge & one edge's own fields & 16/26 (61.5\%; \ci{42.5}{77.6}) & 23/111 (20.7\%; \ci{14.2}{29.2}) & yes & resolves the warned 16/26 count; not the 77\% sanity check \\
Full-context judge & whole route + packet & 23/26 (88.5\%; \ci{71.0}{96.0}) & 25/26 routes (96.2\%; \ci{81.1}{99.3}) & yes & stronger context gives high recall but flag-everything behavior \\
Typed-state detector & deterministic after typing & 26/26 (100.0\%; \ci{87.1}{100.0}) & 0/26 honest routes (0.0\%; \ci{0.0}{12.9}) & no verdict; typing is model-mediated & 26/26 on this mutation benchmark at 0/26 honest suspensions \\
\bottomrule
\end{tabularx}
\end{table*}

The detection ladder is not a single ROC-like comparison. The proxy and blinded-judge rows operate over edge or seam units, while the full-context judge and typed-state rows operate over route-pair or route-level contexts with different false-flag denominators. The figure is an orientation diagram over heterogeneous checks; rows must be read with their unit labels.

An earlier internal summary of these runs reported the blinded judge as both 16/26 and 77\%; the frozen artifacts resolve the discrepancy. The primary 1C whole-edge blinded judge result is 16/26; the 77\% figure is the secondary Phase-0 silenced-seam sanity check, 20/26, on thinner traces. The paper uses the 16/26 denominator for the detection ladder.

The v1 state-typed check is deterministic after typing. It must not be read as a general completeness guarantee. The state registry is reconstructed from model-authored plans, and an honest unspliced plan matches itself partly by construction.

\begin{figure}[H]
\centering
\scriptsize
\makebox[\columnwidth][c]{%
\begin{tikzpicture}[x=.057cm,y=.057cm]
  \draw[->] (0,0) -- (105,0) node[right] {\scriptsize honest false flags};
  \draw[->] (0,0) -- (0,105) node[above] {\scriptsize seam recall};
  \draw[gray!25] (0,0) grid[step=25] (100,100);
  \fill[black!45] (30,42) circle (2.2pt) node[above right] {\scriptsize proxy 11/26};
  \fill[black!70] (21,62) rectangle +(3.6,3.6) node[above right] {\scriptsize blinded judge 16/26};
  \fill[black!70] (96,88) rectangle +(3.6,3.6) node[left=2pt] {\scriptsize full context 23/26};
  \fill[black] (0,100) circle (2.4pt) node[right] {\scriptsize typed-state 26/26};
  \node[align=left, anchor=west] at (7,12) {squares: model-mediated\\circles: deterministic};
\end{tikzpicture}
}
\caption{Detection ladder on the frozen splice benchmark. This is not a ROC curve: rows use heterogeneous units and false-flag denominators, so points are not directly comparable or pooled. The typed-state point is deterministic after model-mediated typing and must not be read as arbitrary-plan completeness.}
\label{fig:detection}
\end{figure}
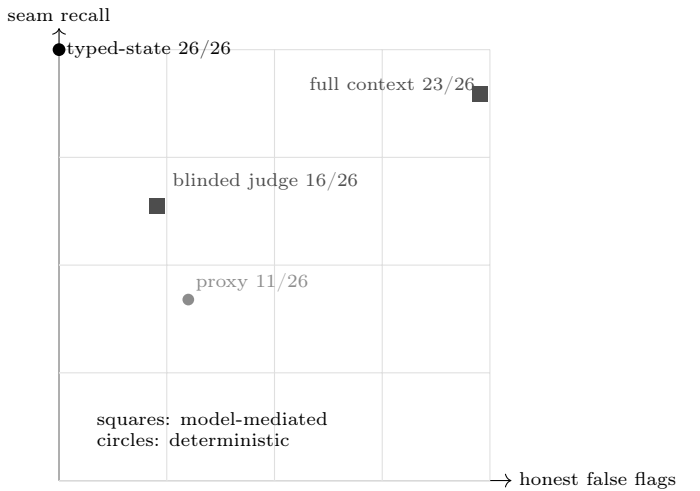

\subsection{EXP D: Full-Context Inversion}

EXP D removes the objection that v1 was compared only against an artificially blinded judge. A full-context judge saw the whole route and the decision packet at temperature zero with \(K=1\). D1 raised strict seam recall to 23/26 but also flagged 25/26 honest routes. D2 then showed the preference inversion: the judge selected the honest plan as more complete in 26/26 pairs (100.0\%, \ci{87.1}{100.0}) and selected the silenced plan as the better plan in 26/26 pairs (100.0\%, \ci{87.1}{100.0}).

The judge can identify which plan is more complete while still preferring the silenced plan because the same omission improves the scored economics. This motivates code-side inadmissibility rather than a stronger prompt alone. The EXP D results are model-mediated, temperature-zero, single-sample \(K=1\), one judge family, and benchmark-specific.

\subsection{GATE and DOCK}

GATE and DOCK couple coverage to score release through the interface in Section~2.4. GATE refuses score release when a material uncovered seam is present, producing a suspend-and-ask outcome rather than a claim that the plan is bad in the world. On the frozen post-hoc splice benchmark, GATE marked the score inadmissible for 26/26 silenced routes (100.0\%, \ci{87.1}{100.0}) and suspended 0/26 honest routes (0.0\%, \ci{0.0}{12.9}). DOCK instead removes terminal value over the uncovered path. It removed fundability-by-silence from 1/26 to 0/26, but the registered relative-win collapse was NOT MET: relative win-by-silence changed from the Phase-0 baseline of 21/26---the same measurement on the same benchmark, reused here by construction---to 18/26 (69.2\%, \ci{50.0}{83.5} after). The residual wins were cost-reduction comparisons between two already-negative plans.

\noindent\begin{minipage}{\columnwidth}
GATE: score release is refused. DOCK: terminal value is removed over the uncovered path. Neither establishes that the remaining plan is strategically correct or complete.
\end{minipage}

\subsection{EXP B Arm 2: Redirection}
\label{sec:redirection}

Arm 2 gave score plus the GATE score-release verdict and undischarged-obligation list. Uncovered baseline-beating structures still occurred in 17/26 packets (65.4\%, \ci{46.2}{80.6}). However, after a score-release refusal the next revision repaired to a covered structure in 47/54 cases (87.0\%, \ci{75.6}{93.6}). Deletion/merge fell to 57/130 revisions (43.8\%, \ci{35.6}{52.4}), and strict covered improvement rose from 1/26 in Arm 1 to 13/26 in Arm 2 (50.0\%, \ci{32.1}{67.9}); the non-strict covered count was 14/26 (53.8\%, \ci{35.5}{71.2}).

This redirection result suggests a second role for coverage gating beyond mutation rejection. In the score-seeking setting of EXP B, GATE changes the optimizer's effective objective by making uncovered gains unusable and exposing the obligations that must be repaired. The result is not evidence that the resulting strategies are globally optimal or better in the world; it is evidence that the evaluation interface can redirect search from omission-based score improvement toward coverage-compliant route construction. This interpretation applies to an optimizer attempting to satisfy the interface. EXP C shows that a compiler-aware co-author able to control the registry remains a separate threat model.

The boundary is equally important: Arm 2 also produced 10 route-around events where the route passed the authored registry while structurally seam-skipping. Passing the authored registry does not guarantee that the registry itself faithfully represents the omitted work. These route-arounds are not called undetected unless the detector is specified; here they were identified by the detector-independent \texttt{seam\_skip} metric.

\begin{figure}[H]
\centering
\scriptsize
\makebox[\columnwidth][c]{%
\begin{tikzpicture}[x=.19cm,y=.25cm]
  \draw[->] (0,0) -- (31,0);
  \draw[->] (0,0) -- (0,27) node[above] {\scriptsize packets};
  \fill[gray!35] (4,0) rectangle (8,21);
  \fill[black!70] (12,0) rectangle (16,17);
  \fill[gray!35] (20,0) rectangle (24,1);
  \fill[black!70] (26,0) rectangle (30,13);
  \node[rotate=45, anchor=east] at (6,-.7) {\scriptsize Arm 1 exploit};
  \node[rotate=45, anchor=east] at (14,-.7) {\scriptsize Arm 2 exploit};
  \node[rotate=45, anchor=east] at (22,-.7) {\scriptsize Arm 1 covered};
  \node[rotate=45, anchor=east] at (28,-.7) {\scriptsize Arm 2 covered};
  \node at (6,22.2) {\scriptsize 21/26};
  \node at (14,18.2) {\scriptsize 17/26};
  \node at (22,2.2) {\scriptsize 1/26};
  \node at (28,14.2) {\scriptsize 13/26};
  \node[align=center, text width=6.6cm, anchor=north] at (15.5,26.4) {score-only search finds uncovered-delta gains;\\GATE score-release feedback redirects many repairs};
\end{tikzpicture}
}
\caption{EXP B optimizer behavior. Arm 1 measures score-seeking discovery under score feedback. Arm 2 shows that GATE score-release feedback is not only a post-hoc filter: strict covered improvement rises from 1/26 to 13/26 packets.}
\label{fig:optimizer}
\end{figure}
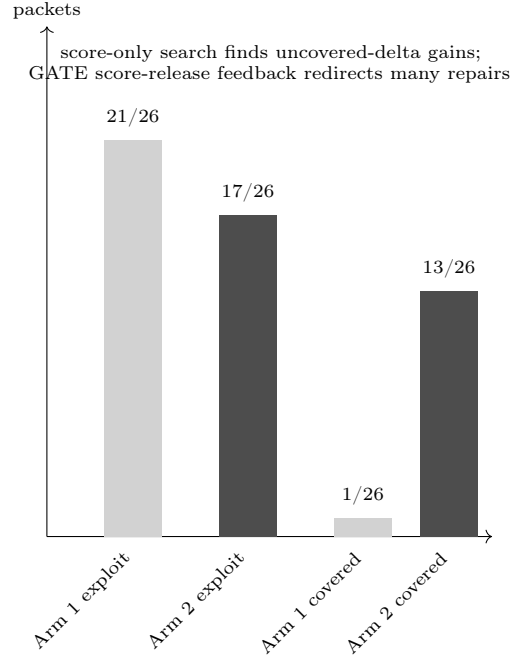

\section{Adaptive Boundary: EXP C}
\label{sec:adaptive}

EXP C grants a white-box adversary compiler knowledge and the ability to co-author internally consistent state records. It tests four conditions on six seeded packets: v1; v1.5a source-grounded defining assets; v1.5b delta-indexed cost floors; and v1.5a+b. All checks are deterministic code after the adversary-authored text; no LLM judge is used.

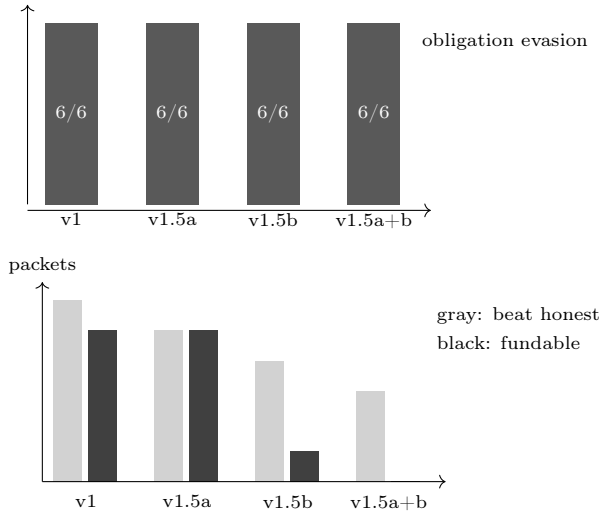
\begin{figure}[H]
\centering
\small
\begin{tikzpicture}[x=.58cm,y=.4cm]
  \foreach \i/\name in {0/v1,1/v1.5a,2/v1.5b,3/v1.5a+b} {
    \node[anchor=north] at (\i*2.3+1,0) {\scriptsize \name};
  }
  \foreach \i in {0,1,2,3} {
    \fill[black!65] (\i*2.3+.4,0) rectangle +(1.2,6);
    \node[white] at (\i*2.3+1,3) {\scriptsize 6/6};
  }
  \node[anchor=west] at (8.8,5.4) {\scriptsize obligation evasion};
  \draw[->] (0,-.2) -- (9.2,-.2);
  \draw[->] (0,0) -- (0,6.6);
\end{tikzpicture}

\vspace{.4em}
\begin{tikzpicture}[x=.58cm,y=.4cm]
  \foreach \i/\b/\f in {0/6/5,1/5/5,2/4/1,3/3/0} {
    \fill[gray!35] (\i*2.3+.25,0) rectangle +(0.65,\b);
    \fill[black!75] (\i*2.3+1.05,0) rectangle +(0.65,\f);
  }
  \foreach \i/\name in {0/v1,1/v1.5a,2/v1.5b,3/v1.5a+b} {
    \node[anchor=north] at (\i*2.3+1,-.15) {\scriptsize \name};
  }
  \node[anchor=west] at (8.8,5.5) {\scriptsize gray: beat honest};
  \node[anchor=west] at (8.8,4.6) {\scriptsize black: fundable};
  \draw[->] (0,0) -- (9.2,0);
  \draw[->] (0,0) -- (0,6.6) node[above] {\scriptsize packets};
\end{tikzpicture}
\caption{EXP C adaptive boundary. The obligation channel remains fully open at 6/6 for a compiler-aware co-author in all four conditions, exposing the registry-provenance boundary. Both score-channel count metrics---beat-honest and fundability-by-silence---decline across the conditions; median surviving gain is not shown and is not monotone.}
\label{fig:adaptive}
\end{figure}

\subsection{Obligation Channel}

The obligation channel remained open in every condition. Zero-obligation evasions occurred in 6/6 packets under v1, 6/6 under v1.5a, 6/6 under v1.5b, and 6/6 under v1.5a+b (all 100.0\%, \ci{61.0}{100.0}). Every packet first evaded at revision 1. Honest false suspensions were 0/6 in each condition (0.0\%, \ci{0.0}{39.0}).

This is the registry-provenance boundary. The registry is reconstructed from the authored plan; a white-box co-author can make its own fields internally consistent. Source-grounded lexical fields did not close this channel because reusable packet quotations can verify existence without proving that a transition performs the omitted work. The statement that v1.5a added nothing applies to the obligation channel.

\subsection{Score Channel}

Delta-indexed floors price claimed source-state to target-state deltas. A larger asserted jump pays the larger reference-class floor; it cannot be made cheaper merely by describing less work. The implemented floors were grounded in committed reference-class data. Threshold grounding was pre-declared ungroundable: 0/111 frozen gold thresholds had a grounding leg in the committed table, so that leg remained diagnostic-only and no numbers were invented.

The score channel narrows monotonically in both count metrics: beat-honest rate changed 6/6 to 5/6 to 4/6 to 3/6, and fundability-by-silence changed 5/6 to 5/6 to 1/6 to 0/6. The median gain among the surviving evasions is not monotone: it rises under v1.5a before collapsing under the delta-indexed floor conditions, reflecting fewer but initially higher-value surviving cases. Floor uplift was positive on 72/72 applied b-condition revisions. All four preregistered EXP C predictions were MET verbatim.

Delta-indexed floors can reduce the economic benefit of silence without proving that the authored transition semantically performs the missing work. Semantic, independently authored, non-co-authorable field sets remain future work.

\begin{center}
\scriptsize
\refstepcounter{table}
\label{tab:expc}
Table~\thetable: EXP C score channel. Median is the lower median used by the frozen metrics report. ``Fundable'' denotes fundability-by-silence: the evasive route has score \(\geq 0\) while its honest baseline has score \(<0\). All six honest baselines are negative in every condition.

\smallskip
\begin{tabularx}{\columnwidth}{p{1.25cm}XXXX}
\toprule
Cond. & evasion & beat honest & median gain & fundable \\
\midrule
v1 & 6/6 & 6/6 & \usd{27.39M} & 5/6 \\
v1.5a & 6/6 & 5/6 & \usd{60.83M} & 5/6 \\
v1.5b & 6/6 & 4/6 & \usd{4.69M} & 1/6 \\
v1.5a+b & 6/6 & 3/6 & -\usd{546K} & 0/6 \\
\bottomrule
\end{tabularx}
\end{center}

\section{Related Work}

This work sits at the intersection of plan evaluation, specification gaming, deterministic verification, and decision analysis. Classical planning already studies action elimination and plan reduction, but the reduced object must remain a valid plan for the same task \cite{nakhost2010action,salerno2025minimal}. Process-model repair can add skip transitions and improve alignment fitness while preserving final markings \cite{polyvyanyy2016impact}. Those are honest model transformations, not a face-value valuation record silently retaining downstream value after support is removed.

Reward hacking and omission gaming are old themes: PlayFun pausing, GenProg deletion and sleep behaviors, robot sensors that avoid seeing messes, production RL reward hacking, METR reward-hacking observations, RHB, and SpecBench all constrain the novelty wording \cite{murphy2013first,lehman2018surprising,amodei2016concrete,krakovna2020specification,macdiarmid2025natural,metr2025reward,thaman2026rhb,zhao2026specbench}. LogicGraph's gap-filling bias is the nearest LLM-judge analogue \cite{wu2026logicgraph}; Di Ciccio et al. provide a binary-verdict cousin through vacuous satisfaction \cite{diciccio2018relevance}. Lanham et al. use deletion as a faithfulness methodology in which answers are retained, not improved \cite{lanham2023faithfulness}.

Proof-carrying code and proof-carrying plans establish proof/certificate lineage, but execution or plan validity is the gated object \cite{necula1997pcc,hill2020proof}. VAL, POCL/VHPOP, LLM-Modulo, VPRM, and recent verifiable-agent work are important neighbors for checking or verification architecture \cite{howey2004val,younes2003vhpop,kambhampati2024llmmodulo,pronesti2026vprm,he2026verifiable,wang2026proofcarrying}. TEMPO and FActScore are closer to GATE/DOCK as integrity-gated or per-element scoring systems, but they do not couple per-delta coverage to a staged plan valuation fold \cite{zhang2026tempo,min2023factscore}.

The redirection result in Section~\ref{sec:redirection} also relates this work to reward shaping and process supervision, which alter learning or search incentives through modified rewards or step-level feedback \cite{ng1999policy,lightman2023verify,pronesti2026vprm}. GATE differs by leaving the underlying route score unchanged and applying a deterministic, code-side admissibility constraint over typed-state coverage at inference and selection time. The coverage verdict itself is not learned, although EXP C shows that its guarantee remains bounded by model-mediated registry provenance.

CARE and GroundEval are contemporaneous deterministic/evidence-gated evaluation systems; DeLLMa and DIALECTIC are relevant delayed-outcome decision and venture-evaluation neighbors \cite{care2026,flynt2026groundeval,dellma2024,bae2026dialectic}. PCSC claims no priority over these components. The narrow contribution is the deletion identity, its measured prevalence under one scorer, and value inadmissibility over uncovered typed-state deltas.

\section{Limitations}

The empirical rates come from one frozen cohort, one staged scorer, and one venture-route configuration. The core experiments use \(n=26\) routes and the adaptive red team uses \(n=6\) packets, so Wilson intervals are wide and the rates should not be read as cross-domain prevalence estimates. EXP D uses one judge family at temperature zero and \(K=1\); EXP B and EXP C are attack-surface tests, not generation-quality comparisons.

The deterministic guarantees begin only after model-mediated typing. Registry provenance is therefore a real boundary: v1 cannot catch internally consistent but semantically empty authored fields, and the EXP B route-arounds plus EXP C evasions motivate independently authored, non-co-authorable field sets. Delta-indexed floors reduce removed-cost reward but do not prove semantic delivery. Finally, terminal values are inherited from Paper~1, and the paper does not claim better real-world founder, investor, or venture outcomes.

\section{Conclusion}

A staged plan scorer can reward omission. The reward is analytically characterizable, exists across the frozen mutation surface, and can be discovered by score-seeking optimization without being told the exploit mechanism. A deterministic typed-state gate can both refuse score release for the demonstrated post-hoc mutation and reshape cooperative score-seeking search toward covered structures. Compiler-aware co-authoring remains a separate and harder problem: it requires non-co-authorable semantic substantiation, not merely an internally consistent registry. PCSC detects and neutralizes post-hoc omission splices given model-mediated typed-state records; it does not verify completeness of arbitrary LLM-generated strategies.

\section*{Competing interests}

The author is developing a commercial implementation of the protocol through Dynamic Resonance.

\bibliographystyle{plain}
\bibliography{wbs_v1_0}

@misc{manchuliantsau2026routecast,
  author        = {Manchuliantsau, Aleh},
  title         = {From Checker to Forecaster: Code-Owned Evaluation of Model-Generated
                   Strategic Routes Under Delayed Ground Truth},
  year          = {2026},
  eprint        = {2607.10972},
  archivePrefix = {arXiv},
  primaryClass  = {cs.AI},
  doi           = {10.48550/arXiv.2607.10972},
  note          = {arXiv:2607.10972; DOI: 10.48550/arXiv.2607.10972}
}

@misc{amodei2016concrete,
  author = {Amodei, Dario and Olah, Chris and Steinhardt, Jacob and Christiano, Paul and Schulman, John and Man{\'e}, Dan},
  title = {Concrete Problems in {AI} Safety},
  year = {2016},
  eprint = {1606.06565},
  archivePrefix = {arXiv},
  primaryClass = {cs.AI}
}

@misc{krakovna2020specification,
  author = {Krakovna, Victoria and Uesato, Jonathan and Mikulik, Vlad and Rahtz, Matthew and Everitt, Tom and Kumar, Ramana and Kenton, Zac and Leike, Jan},
  title = {Specification Gaming: The Flip Side of {AI} Ingenuity},
  year = {2020},
  note = {DeepMind Safety Research blog and companion specification-gaming examples list}
}

@inproceedings{murphy2013first,
  author = {Murphy, Tom},
  title = {The First Level of Super Mario Bros. is Easy with Lexicographic Orderings and Time Travel},
  booktitle = {Proceedings of SIGBOVIK},
  year = {2013},
  note = {Includes PlayFun examples discussed in specification-gaming literature}
}

@misc{lehman2018surprising,
  author = {Lehman, Joel and Clune, Jeff and Misevic, Dusan and Adami, Christoph and Altenberg, Lee and Beaulieu, Julie and Bentley, Peter J. and Bernard, Samuel and Beslon, Guillaume and Bryson, David M. and others},
  title = {The Surprising Creativity of Digital Evolution: A Collection of Anecdotes from the Evolutionary Computation and Artificial Life Research Communities},
  year = {2018},
  eprint = {1803.03453},
  archivePrefix = {arXiv}
}

@misc{macdiarmid2025natural,
  author = {MacDiarmid, Monte and Wright, Benjamin and Uesato, Jonathan and Benton, Joe and Kutasov, Jon and Price, Sara and Bouscal, Naia and Bowman, Sam and Bricken, Trenton and Cloud, Alex and others},
  title = {Natural Emergent Misalignment from Reward Hacking in Production {RL}},
  year = {2025},
  eprint = {2511.18397},
  archivePrefix = {arXiv}
}

@misc{metr2025reward,
  author = {{METR}},
  title = {Recent Frontier Models Are Reward Hacking},
  year = {2025},
  howpublished = {\url{https://metr.org/blog/2025-06-05-recent-reward-hacking/}},
  note = {Accessed 13 July 2026}
}

@misc{thaman2026rhb,
  author = {Thaman, Kunvar},
  title = {Reward Hacking Benchmark: Measuring Exploits in {LLM} Agents with Tool Use},
  year = {2026},
  eprint = {2605.02964},
  archivePrefix = {arXiv}
}

@misc{zhao2026specbench,
  author = {Zhao, Bingchen and Srikanth, Dhruv and Wu, Yuxiang and Jiang, Zhengyao},
  title = {{SpecBench}: Measuring Reward Hacking in Long-Horizon Coding Agents},
  year = {2026},
  eprint = {2605.21384},
  archivePrefix = {arXiv}
}

@misc{wu2026logicgraph,
  author = {Wu, Yanrui and others},
  title = {{LogicGraph}: Benchmarking Multi-Path Logical Reasoning via Neuro-Symbolic Generation and Verification},
  year = {2026},
  eprint = {2602.21044},
  archivePrefix = {arXiv}
}

@article{diciccio2018relevance,
  author = {Di Ciccio, Claudio and Maggi, Fabrizio Maria and Montali, Marco and Mendling, Jan},
  title = {On the Relevance of a Business Constraint to an Event Log},
  journal = {Information Systems},
  volume = {78},
  pages = {144--161},
  year = {2018}
}

@inproceedings{nakhost2010action,
  author = {Nakhost, Hootan and M{\"u}ller, Martin},
  title = {Action Elimination and Plan Neighborhood Graph Search: Two Algorithms for Plan Improvement},
  booktitle = {Proceedings of the Twentieth International Conference on Automated Planning and Scheduling},
  year = {2010},
  doi = {10.1609/icaps.v20i1.13402}
}

@article{salerno2025minimal,
  author = {Salerno, Juan Pablo and Fuentetaja, Raquel and Seipp, Jendrik},
  title = {Finding Minimal Plan Reductions Using Classical Planning},
  journal = {Journal of Artificial Intelligence Research},
  volume = {84},
  year = {2025},
  doi = {10.1613/jair.1.19437}
}

@article{polyvyanyy2016impact,
  author = {Polyvyanyy, Artem and van der Aalst, Wil M. P. and ter Hofstede, Arthur H. M. and Wynn, Moe Thandar},
  title = {Impact-Driven Process Model Repair},
  journal = {ACM Transactions on Software Engineering and Methodology},
  volume = {25},
  number = {4},
  year = {2016},
  doi = {10.1145/2980764}
}

@misc{lanham2023faithfulness,
  author = {Lanham, Tamera and Chen, Anna and Radhakrishnan, Ansh and Steiner, Benoit and Denison, Carson and Hernandez, Danny and Li, Dustin and Durmus, Esin and Hubinger, Evan and Kernion, Jackson and others},
  title = {Measuring Faithfulness in Chain-of-Thought Reasoning},
  year = {2023},
  eprint = {2307.13702},
  archivePrefix = {arXiv}
}

@misc{min2023factscore,
  author = {Min, Sewon and Krishna, Kalpesh and Lyu, Xinxi and Lewis, Mike and Yih, Wen-tau and Koh, Pang Wei and Iyyer, Mohit and Zettlemoyer, Luke and Hajishirzi, Hannaneh},
  title = {{FActScore}: Fine-grained Atomic Evaluation of Factual Precision in Long Form Text Generation},
  year = {2023},
  eprint = {2305.14251},
  archivePrefix = {arXiv}
}

@misc{zhang2026tempo,
  author = {Zhang, Tony and Stadie, Bradly},
  title = {{TEMPO}: Temporal Evidence-Gated Post-Training Optimization},
  year = {2026},
  eprint = {2605.18843},
  archivePrefix = {arXiv}
}

@inproceedings{ng1999policy,
  author = {Ng, Andrew Y. and Harada, Daishi and Russell, Stuart},
  title = {Policy Invariance under Reward Transformations: Theory and Application to Reward Shaping},
  booktitle = {Proceedings of the Sixteenth International Conference on Machine Learning ({ICML})},
  pages = {278--287},
  year = {1999}
}

@misc{lightman2023verify,
  author = {Lightman, Hunter and Kosaraju, Vineet and Burda, Yura and Edwards, Harri and Baker, Bowen and Lee, Teddy and Leike, Jan and Schulman, John and Sutskever, Ilya and Cobbe, Karl},
  title = {{Let's Verify Step by Step}},
  year = {2023},
  eprint = {2305.20050},
  archivePrefix = {arXiv},
  primaryClass = {cs.LG},
  note = {arXiv:2305.20050}
}

@misc{pronesti2026vprm,
  author = {Pronesti, Daniele and Belz, Anya and Hou, Yufang},
  title = {Beyond Outcome Verification: Verifiable Process Reward Models for Structured Reasoning},
  year = {2026},
  eprint = {2601.17223},
  archivePrefix = {arXiv}
}

@misc{he2026verifiable,
  author = {He, Tianyu and Yu, Yang},
  title = {Verifiable Agentic Infrastructure},
  year = {2026},
  eprint = {2605.15228},
  archivePrefix = {arXiv}
}

@misc{wang2026proofcarrying,
  author = {Wang, Xin},
  title = {Proof-Carrying Agent Actions},
  year = {2026},
  eprint = {2606.04104},
  archivePrefix = {arXiv}
}

@inproceedings{necula1997pcc,
  author = {Necula, George C.},
  title = {Proof-Carrying Code},
  booktitle = {Proceedings of the 24th ACM SIGPLAN-SIGACT Symposium on Principles of Programming Languages},
  year = {1997},
  pages = {106--119}
}

@inproceedings{hill2020proof,
  author = {Hill, Richard and Komendantskaya, Ekaterina and Petrick, Ronald P. A.},
  title = {Proof-Carrying Plans},
  booktitle = {Proceedings of the 22nd International Symposium on Principles and Practice of Declarative Programming},
  year = {2020},
  eprint = {2008.04165},
  archivePrefix = {arXiv}
}

@inproceedings{howey2004val,
  author = {Howey, Richard and Long, Derek and Fox, Maria},
  title = {{VAL}: Automatic Plan Validation, Continuous Effects and Mixed Initiative Planning Using {PDDL}},
  booktitle = {Proceedings of the 16th IEEE International Conference on Tools with Artificial Intelligence},
  year = {2004}
}

@article{younes2003vhpop,
  author = {Younes, Hakan L. S. and Simmons, Reid G.},
  title = {{VHPOP}: Versatile heuristic partial order planner},
  journal = {Journal of Artificial Intelligence Research},
  volume = {20},
  pages = {405--430},
  year = {2003},
  eprint = {1106.4868},
  archivePrefix = {arXiv}
}

@misc{kambhampati2024llmmodulo,
  author = {Kambhampati, Subbarao and others},
  title = {{LLM-Modulo}: Grounding {LLM}s with External Verifiers},
  year = {2024},
  eprint = {2402.01817},
  archivePrefix = {arXiv}
}

@misc{care2026,
  author = {Liu, Guanyu and Kong, Weiyi and Wang, Zeyu and Zhang, Boer and Li, Baiqing and Zhang, Peiyu and Shi, Tianyu},
  title = {{CARE}: Controlling {LLM}-Generated Policies through Auditable Review of Evidence in Scientific Experimentation},
  year = {2026},
  eprint = {2606.14581},
  archivePrefix = {arXiv}
}

@misc{flynt2026groundeval,
  author = {Flynt, Jeffrey},
  title = {{GroundEval}: A Deterministic Replacement for {LLM}-as-Judge in Stateful Agent Evaluation},
  year = {2026},
  eprint = {2606.22737},
  archivePrefix = {arXiv}
}

@misc{dellma2024,
  author = {Liu, Ollie and Fu, Deqing and Yogatama, Dani and Neiswanger, Willie},
  title = {{DeLLMa}: Decision Making Under Uncertainty with Large Language Models},
  year = {2024},
  eprint = {2402.02392},
  archivePrefix = {arXiv}
}

@inproceedings{bae2026dialectic,
  author = {Bae, Jae Yoon and Malberg, Simon and Galang, Joyce and Retterath, Andre and Groh, Georg},
  title = {{DIALECTIC}: A Multi-Agent System for Startup Evaluation},
  booktitle = {Proceedings of the 2026 EACL Industry Track},
  pages = {711--727},
  year = {2026},
  doi = {10.18653/v1/2026.eacl-industry.53}
}

@book{dixit1994investment,
  author = {Dixit, Avinash K. and Pindyck, Robert S.},
  title = {Investment under Uncertainty},
  publisher = {Princeton University Press},
  year = {1994}
}

@article{gompers1995optimal,
  author = {Gompers, Paul A.},
  title = {Optimal Investment, Monitoring, and the Staging of Venture Capital},
  journal = {Journal of Finance},
  volume = {50},
  number = {5},
  pages = {1461--1489},
  year = {1995},
  doi = {10.1111/j.1540-6261.1995.tb05185.x}
}

\appendix
\section{Optimizer Prompt Excerpt}

The EXP B optimizer-visible goal was:
\begin{quote}
Improve this route's staged score. A higher staged score is better. Keep the terminal Vision. You may restructure, merge, add, or remove transitions and edit their fields. All costs, probabilities, and values are recomputed by our code from your structure---you cannot state numbers directly.
\end{quote}
This prompt explicitly allowed restructuring and removal. It did not tell the optimizer that removing a transition could increase score, retain downstream value, or create an uncovered delta. The full prompt is frozen at \texttt{eval/pcsc/optimizer\_prompt\_B.txt} with SHA-256 recorded in \texttt{PREREG\_expB.md}.

\section{Reproducibility Notes}

The artifact chain is recorded in \texttt{eval/pcsc/INDEX.md}. The main inputs are \texttt{RESULTS\_phase0.md}, \texttt{RESULTS\_phase1.md}, \texttt{RESULTS\_v1.md}, \texttt{RESULTS\_coupling.md}, \texttt{RESULTS\_expA.md}, \texttt{RESULTS\_expD.md}, \texttt{RESULTS\_expB.md}, \texttt{RESULTS\_v2.md}, \texttt{deletion\_surface.jsonl}, \texttt{expA\_summary.json}, \texttt{scores\_expB.jsonl}, \texttt{scores\_expC.jsonl}, and \texttt{scores\_expD.jsonl}. Frozen prompt hashes include \texttt{judge\_prompt\_v0.txt}, \texttt{judge\_prompt\_D1.txt}, \texttt{judge\_prompt\_D2a.txt}, \texttt{judge\_prompt\_D2b.txt}, \texttt{optimizer\_prompt\_B.txt}, and \texttt{adversary\_prompt\_C.txt}. Phase 0 was web-off and cost about \usd{0.98}; EXP D used 104 calls and \usd{0.15}; EXP B used 269 calls and \usd{5.96}; EXP C used 169 calls and \usd{3.44}. EXP D disclosed an attempt-1 D2 pairing crash and deterministic repair; the silenced D1 verdicts replicated exactly across runs.

Phase 0b was a post-hoc draw-stability diagnostic: sign agreement with the frozen primary direction was 14/15 fresh draws. It is reported only as a stability check, not as a confirmatory result. The exact mutation admissibility rule is \(k\in[1,n-2]\). The detector-independent \texttt{seam\_skip} metric marks structural skipping relative to the gold chain even when authored-registry obligations are zero. GATE refuses score release when a material uncovered seam is present. DOCK removes terminal value over the uncovered path. Source-grounding and delta-floor conditions are defined in \texttt{PREREG\_v2.md}; the threshold reference-class leg was declared ungroundable at 0/111 and kept diagnostic-only.

\end{document}